\pdfoutput=1

\documentclass[11pt]{article}

\usepackage[final]{EACL2023}

\usepackage{times}
\usepackage{latexsym}
\usepackage{array}
\usepackage{multirow}
\usepackage{color}
\usepackage{pgfplots}
\usepackage{dingbat}
\usepackage{pifont}
\usepackage{contour}
\usepackage{cancel}
\usepackage[normalem]{ulem}
\usepackage{dblfloatfix}

\contourlength{0.8pt}

\newcommand{\myuline}[1]{%
  \uline{\phantom{#1}}%
  \llap{\contour{white}{#1}}%
}

\newcommand{\xmark}{\ding{55}}%

\usepackage[T1]{fontenc}

\usepackage[utf8]{inputenc}

\usepackage{microtype}

\usepackage{inconsolata}

\title{Better Handling Coreference Resolution in Aspect Level Sentiment Classification by Fine-Tuning Language Models}

\author{Dhruv Mullick \\
  Dept. of Computing Science \\
  University of Alberta \\
  \texttt{mullick@ualberta.ca} \\\And
  Bilal Ghanem \\
  Dept. of Computing Science \\
  University of Alberta \\
  \texttt{bilalhgm@gmail.com} \\\\\And
  Alona Fyshe \\
  Dept. of Computing Science \\
  University of Alberta \\
  \texttt{alona@ualberta.ca} \\
  }

\begin{document}
\maketitle
\begin{abstract}
Customer feedback is invaluable to companies as they refine their products. Monitoring customer feedback can be automated with Aspect Level Sentiment Classification (ALSC) which allows us to analyse specific aspects of the products in reviews. Large Language Models (LLMs) are the heart of many state-of-the-art ALSC solutions, but they perform poorly in some scenarios requiring Coreference Resolution (CR). In this work, we propose a framework to improve an LLM's performance on CR-containing reviews by fine tuning on highly inferential tasks. We show that the performance improvement is likely attributed to the improved model CR ability. We also release a new dataset that focuses on CR in ALSC.

\end{abstract}

\section{Introduction}

To understand an end user's perspective on a product, it is common to consider reviews on online platforms. A company can look for the customers' perspective on a certain aspect of the product. For instance, a laptop company might look for reviews concerning "battery." Aspect Level Sentiment Classification (ALSC) analyzes reviews for sentiments of specific aspects, like the "battery" aspect in earlier example \cite{yan-etal-2021-unified}. ALSC is a sub-task of a wider body of work called Aspect Based Sentiment Analysis (ABSA) \cite{liu_sentiment_2012}, which aims to extract aspects and their associated sentiments. State-of-the-art ALSC solutions often use Large Language Models (LLMs) \cite{zhang2022survey}. 

Reviews often use pronouns, which can make coreference resolution (CR) in LLMs necessary to infer the sentiment associated with the aspect. Hence, LLMs used for ALSC need strong CR ability, and can fail otherwise. For instance, the sentence - "\textit{He ate food at the restaurant, it was deserted.}" requires the LLM to understand that the definite pronoun "it" refers to the "restaurant" (antecedent), because of the context ("deserted"). Table \ref{table:cr_cases} shows four examples where the state-of-the-art T5 ALSC model \cite{zhang-etal-2021-towards-generative} fails due to its poor CR ability. We find that \textasciitilde 15\% of this T5 model's errors are on cases requiring CR ability.

\begin{table}[!tb]
\centering
\caption{Cases where the T5 ALSC model fails due to its poor coreference resolution ability.}
\label{table:cr_cases}
\small
\begin{tabular}{|l|l|ll|}
\hline
\textbf{Sentence} & \textbf{Aspect} & \multicolumn{2}{c|}{\textbf{Sentiment Polarity}} \\
\textbf{} & \textbf{} & \multicolumn{1}{l|}{\textbf{Predicted}} & \textbf{Gold} \\ \hline
\multirow{2}{*}{\parbox{3cm}{\strut He ate food at the restaurant, it was \underline{deserted} \strut}} & restaurant & \multicolumn{1}{l|}{neutral} & negative \\ \cline{2-4} 
 & food & \multicolumn{1}{l|}{negative} & neutral \\ \hline
\multirow{2}{*}{\parbox{3cm}{\strut He ate food at the restaurant, it was \underline{dark} \strut}} & restaurant & \multicolumn{1}{l|}{neutral} & negative \\ \cline{2-4} 
 & food & \multicolumn{1}{l|}{negative} & neutral \\ \hline
\end{tabular}
\end{table}

LLMs are also known to have performance and stability issues \cite{phang2018sentence}. To remedy these, instead of directly training on the task of interest (target task), it can be beneficial to first train on an auxiliary task \cite{pruksachatkun-etal-2020-intermediate}. Certain auxiliary tasks can contribute to both improved performance and stability of the target task \cite{phang2018sentence}. Using auxiliary training, our work shows a way to improve an LLM's performance on English ALSC reviews requiring CR.

In our work, we: \textbf{a)} show that an LLM trained for ALSC makes more errors when evaluated only on reviews requiring CR ability, compared to when handling typical ALSC reviews (8.7\% mean F1); \textbf{b)}~demonstrate that our framework for handling CR-containing reviews can improve ALSC model's CR ability (16\% mean F1); \textbf{c)} show that this improved CR ability can improve ALSC performance for reviews requiring CR ability (5\% mean F1). \textbf{d)} release annotated variants of existing datasets which can be used to benchmark a model's ALSC performance on CR cases.

\section{Experimental Setup}


\subsection{Data}

\paragraph{Original ALSC Datasets} We consider English ALSC datasets: SemEval Restaurant (Rest16) \cite{pontiki-etal-2016-semeval} and MAMS \cite{jiang-etal-2019-challenge}, both of which contain reviews from a similar restaurant domain. Inspired by \citet{yan-etal-2021-unified}, ALSC reviews are processed into an input format suitable for our LLM - "[sentence]. aspect: [aspect]". The ground truth output is "positive", "negative" or "neutral". For example, "\textit{\$20 for good sushi cannot be beaten. aspect: sushi}" has the ground truth as "positive". We clean datasets as per Appendix \ref{dataset_cleanup}.

\paragraph{CR Cases} 

We identify reviews in the Rest16 and MAMS datasets that contain definite pronouns, and henceforth call these sentences \emph{Pronoun cases}.

Limiting ourselves to the ALSC task described above, we say that a review is a \emph{CR case} if its sentiment requires proper coreference resolution for correct classification. Specifically, \emph{the aspect should be an antecedent of a definite pronoun which is associated with a sentiment polarity}. For example, "He ate food at the restaurant, it was deserted." with aspect: "restaurant" is a CR case.  Here, "restaurant" is the antecedent of "it" which is associated with "deserted" and has negative connotations. CR cases are manually selected from Pronoun cases.

\paragraph{ALSC-CR Dataset} Our dataset is composed of the original ALSC datasets (Rest16 and MAMS). The testing, however, is done only using CR cases, and we use a combination of Pronoun and Non-Pronoun cases for validation and train sets. Table~\ref{table:absa_cr_data} presents the dataset composition. Better performance on the test dataset will indicate a superior ability to handle CR cases in ALSC.

The train, validation and test sets are of similar, but not identical, distributions. Due to the limited number of CR cases, it is not possible to have train and validation sets composed entirely of CR cases. More details can be found in Appendix \ref{app:dataset_details}.

\begin{table*}[!h]
\centering
\small
\caption{ALSC-CR composition. Note that CR cases are types of Pronoun cases.}
\label{table:absa_cr_data}
\begin{tabular}{|l|c|c|c|c|c|c|}
\hline
\textbf{\myuline{Partition}} & \textbf{\myuline{Size}} & \multicolumn{2}{c|}{\textbf{\myuline{Dataset}}} & \multicolumn{3}{c|}{\textbf{\myuline{Data Type}}} \\
 &  &  &  & \multicolumn{2}{c|}{\textbf{\textit{Pronoun Cases}}} & \textbf{\textit{Non-Pronoun Cases}} \\
\textbf{} & \textbf{} & \textbf{MAMS} & \textbf{Rest16} & \textbf{CR Cases} & \textbf{Non-CR Pronoun Cases} & \\ \hline
Train & 12,434 & \checkmark & \checkmark & \checkmark & \checkmark & \checkmark \\ \hline
Validation & 889 & \checkmark & \checkmark & \checkmark & \checkmark & \checkmark \\ \hline
Test & 346 & \checkmark & \checkmark & \checkmark & \xmark & \xmark \\ \hline
\end{tabular}
\end{table*}

\subsection{Auxiliary Tasks} \label{aux_tasks}

We use highly inferential tasks for auxiliary training in our experiments as they generally provide higher improvements for various NLP target tasks \cite{pruksachatkun-etal-2020-intermediate}. We select two commonsense tasks - Commongen \cite{lin-etal-2020-commongen} and CosmosQA \cite{huang-etal-2019-cosmos}, as commonsense reasoning helps with CR \cite{liu2017combing}. SQuAD \cite{rajpurkar-etal-2016-SQuAD} is selected because it is a non-commonsense question answering (QA) task. Its performance is contrasted with CosmosQA, checking if it is the QA or the commonsense ability which improves CR. Quora Question Prediction \cite{wang-etal-2018-glue} (QQP) is selected as it benefits performance on the Stanford Sentiment Treebank (SST) task which is similar to ALSC \cite{wang-etal-2019-tell}. Even if auxiliary tasks aren't designed for CR, they can impart CR ability to the model. For the QA example - “Context: Alice can't come. She is old”; “Question: Who is old?”, answer is “Alice”. Answering this requires CR and teaches the model CR ability.

\textbf{Commongen} is a generative commonsense task involving generation of a plausible sentence given a list of concepts (train size = 67,389). It tests: 1) relational reasoning which is the ability to construct grammatical sentences adhering to commonsense; 2) compositional generalisation which is reasoning with unseen concept combinations. For example: input - "concepts = [dog, frisbee, catch, throw]"; output - "A dog leaps to catch a thrown frisbee." 

\textbf{CosmosQA} is a QA task where answering questions requires commonsense (train size = 25,262). For each question, there are four options, and the model should output the correct option number.

\textbf{SQuAD} is an extractive QA task where the correct answer to the question is present exactly in the passage (train size = 87,599). 

\textbf{QQP} task involves checking if two Quora questions are semantically equivalent. We cap the train size at 50,000 to match the other datasets.

\section{Experiments and Results}

We ran experiments for three purposes: \textbf{a)}~to show there is drop in ALSC performance for reviews requiring CR ability; \textbf{b)}~to show we can alleviate this performance drop by auxiliary fine-tuning; \textbf{c)}~to provide additional evidence that change in performance on CR cases is due to improved CR ability. 

Inspired by state-of-the-art performance in \citet{zhang-etal-2021-towards-generative}, we used the T5 LLM \cite{raffel2019exploring}. Our baseline model is a T5 trained on ALSC-CR, but not fine-tuned on auxiliary tasks.

The T5 model was trained in various settings using training prompts / input prefixes (Appendix \ref{app:training_prompts}). Wording of prompts has limited impact on the outcome so we did not experiment with the wording \cite{raffel2019exploring}. Rather, we relied on prior work for task prompts \cite{lin-etal-2020-commongen, Lourie2021UNICORNOR, raffel2019exploring}. For ALSC and Definite Pronoun Resolution (DPR) \cite{rahman-ng-2012-resolving} (Sec. \ref{exp:dpr_cr}), we created prompts as we did not find examples in prior work (see Appendix \ref{app:training_prompts}).

All experiments were run with at least 10 random seeds, and Yuen-Welch test was used for testing statistical significance.

\subsection{Model Performance on ALSC Without Auxiliary Fine Tuning} \label{exp:finetune_absa_reg}

To check LLM performance on CR cases, we evaluated the T5 model on regular ALSC data (ALSC-Regular), which does not consist solely of CR cases. ALSC-Regular and ALSC-CR are equal sized and have an identical proportion of Rest16 and MAMS. We also evaluated the T5 model on ALSC-CR, to get the model's performance solely on CR cases.

By comparing T5 model's performance on the two ALSC datasets, we show that unspecialized LLMs face a significant performance problem while handling reviews requiring CR ability. Results are shown in Table \ref{table:absa_problem}, where evaluation on ALSC-CR shows a drop in performance of \textasciitilde8.7\% mean F1, as well as an increase of 0.6 F1 standard deviation indicating a poorer model convergence.

\begin{table}[!h]
\centering
\small
\caption{T5 model evaluated on ALSC datasets. Best score bolded. Performances on the datasets are statistically significantly different (p-value=$9.03 e-05$).}
\label{table:absa_problem}
\begin{tabular}{|l|c|}
\hline
\textbf{Dataset} & \textbf{Mean F1 {\footnotesize ($\pm$ Std. Dev)}} \\ \hline
ALSC-Regular & \textbf{79.71} {\footnotesize ($\pm$ 1.99)} \\ \hline
ALSC-CR & 71.07 {\footnotesize ($\pm$ 2.60)} \\ \hline
\end{tabular}
\end{table}

\subsection{Fine Tuning With Auxiliary Tasks} \label{exp:finetune_aux}

As a solution to poor performance on ALSC-CR (Section \ref{exp:finetune_absa_reg}), we experimented with various auxiliary tasks mentioned in Section \ref{aux_tasks}.

We trained T5 model on the auxiliary task first to incorporate auxiliary task knowledge. This model is then trained and evaluated on ALSC-CR, our target task. We experimented with different auxiliary dataset sizes as the size has little correlation with the target task performance \cite{wang-etal-2019-tell}.

The model's performance on ALSC-CR with different auxiliary tasks is compared to baseline model's ALSC-CR performance to see if auxiliary tasks were beneficial. Results are shown in Table \ref{table:absa_cr_finetune}. We find that the lower ALSC-CR performance (compared to ALSC-Regular) can be alleviated by auxiliary training with Commongen and QQP, which lead to statistically significant improvements of \textasciitilde 5\% mean F1. Auxiliary training with CosmosQA and SQuAD does not lead to statistically significant improvement in any case.

\begin{table*}[!h]
\centering
\small
\caption{Mean F1 {\footnotesize ($\pm$ Std. Dev)} performance on ALSC-CR on different fractions of aux dataset. * denotes statistically significant difference from baseline. Table's best scores bolded, $2^{nd}$ best underlined.}
\label{table:absa_cr_finetune}
\begin{tabular}{|l|c|c|c|c|}
\hline
\textbf{Aux. Task} & \multicolumn{4}{c|}{\textbf{Aux. Dataset Fraction}} \\
\textbf{} & \multicolumn{1}{c|}{\textbf{0.1}} & \multicolumn{1}{c|}{\textbf{0.2}} & \multicolumn{1}{c|}{\textbf{0.5}} & \multicolumn{1}{c|}{\textbf{1.0}} \\ \hline
Commongen & \myuline{75.72} {\footnotesize ($\pm$ 1.14)} * & 72.46 {\footnotesize ($\pm$ 2.21)} & 71.04 {\footnotesize ($\pm$ 3.50)} & 71.45 {\footnotesize ($\pm$ 1.91)} \\ \hline
CosmosQA & 71.79 {\footnotesize ($\pm$ 1.55)} & 71.45 {\footnotesize ($\pm$ 3.02)} & 72.60 {\footnotesize ($\pm$ 1.85)} & 73.12 {\footnotesize ($\pm$ 2.15)} \\ \hline
SQuAD & 72.02 {\footnotesize ($\pm$ 1.88)} & 72.60 {\footnotesize ($\pm$ 2.07)} & 71.47 {\footnotesize ($\pm$ 3.24)} & 72.08 {\footnotesize ($\pm$ 2.25)}\\ \hline
QQP & 72.49 {\footnotesize ($\pm$ 2.79)} & 71.85 {\footnotesize ($\pm$ 2.98)} & \textbf{76.10} {\footnotesize ($\pm$ 1.26)} * & 71.30 {\footnotesize ($\pm$ 2.19)}\\ \hline
N/A (Baseline) & \multicolumn{4}{c|}{71.07 {\footnotesize ($\pm$ 2.60)}} \\ \hline
\end{tabular}
\end{table*}

Prior work \cite{pruksachatkun-etal-2020-intermediate} showed a general improvement in a model's target task performance when fine-tuned with highly inferential tasks. Apart from being highly inferential, because Commongen is a generative commonsense task, it is ideal for imparting commonsense knowledge to a generative LLM like T5. On the other hand, CosmosQA being a discriminative task is unlikely to impart as much commonsense knowledge into a generative system \cite{lin-etal-2020-commongen}. As being highly inferential is helpful for target tasks, the SQuAD extractive QA task, would not result in as significant an improvement. When used for auxiliary training, QQP shows a high improvement in the SST target task \cite{wang-etal-2019-tell} which involves similar sentiment analysis, explaining QQP's improved performance on ALSC-CR.

Though auxiliary training on DPR might seem promising, it is a much smaller dataset (train size = 1500) than other tasks.  For completeness we did train using DPR but found that the mean F1 = 72.77 was not statistically significantly different from the baseline.

Similar to \citet{wang-etal-2019-tell}, we do not find correlation between auxiliary task size and target performance. This lack of correlation can be since small datasets might not teach the task sufficiently \cite{raffel2019exploring}. On the other hand, large auxiliary datasets can cause catastrophic forgetting of the LLM's original objective \cite{wang-etal-2019-tell}. This original objective is generally beneficial for target tasks. Despite this lack of correlation, we have demonstrated a framework for improving any target task's performance on CR cases.

We show a pronoun error analysis in Appendix \ref{app:absa_cr_pronoun_ea} to better understand the ALSC-CR improvements.

\subsection{Evaluating Coreference Ability} \label{exp:dpr_cr}

Performing well on ALSC-CR requires strong CR ability, as CR associates the aspect with its sentiment. To verify that the improvement in Section \ref{exp:finetune_aux} is attributable to the ALSC model's improved CR ability, we estimate the CR ability by evaluating on DPR. Since we have an ALSC model for each random seed used for training (Section \ref{exp:finetune_aux}), we run DPR evaluation on the ALSC random seed model with the highest ALSC-CR val set performance. 

The DPR task involves predicting the antecedent of the given pronoun. This is precisely the ability required for good performance on ALSC-CR (which contains only definite pronoun cases), making DPR ideal to measure the CR ability of our models. Other CR datasets like OntoNotes \cite{hovy-etal-2006-ontonotes} are not as suitable as DPR because DPR only focuses on definite pronouns, which is the ability we are interested in. Similarly, DPR is also the only CR dataset suitable for auxiliary training, but the size makes this infeasible as discussed in Sec. \ref{exp:finetune_aux}. 

We use a DPR variant for generative models where input is of the form: "Humans were afraid of robots as *they* were strong.", and the objective is to predict what the highlighted pronoun (*they*) is referring to \cite{raffel2019exploring}. 

Evaluating ALSC models on DPR (Table \ref{table:absa_cr_ability}) confirms that the ALSC-CR performance gains may be attributable to the improved CR ability of the model due to auxiliary fine-tuning. Experiments show that Commongen and QQP fine-tuned models show a drastically improved (and statistically significant) CR ability of up to \textasciitilde 16\%. This explains their improved ALSC-CR performance. Using CosmosQA, we see a statistically significant \textasciitilde 5\% deterioration in CR ability which does not lead to statistically significant changes in ALSC-CR performance.

\begin{table}[!h]
\centering
\small
\caption{CR ability of top performing models (Sec \ref{exp:finetune_aux}) measured using DPR. Statistically significant improvement(*) and deterioration(\textsuperscript{\textdagger}) from baseline marked. Best bolded, $2^{nd}$ best underlined.}
\label{table:absa_cr_ability}
\begin{tabular}{|l|c|c|c|}
\hline
\textbf{Aux Task} & \textbf{Aux Frac.} & \textbf{Mean F1 {\footnotesize ($\pm$ Std. Dev)}} \\ \hline
N/A (Baseline) & 0 & 59.28 {\footnotesize ($\pm$ 8.82)} \\ \hline
Commongen & 0.1 & \myuline{75.77} {\footnotesize ($\pm$ 1.68)}* \\ \hline
CosmosQA & 1.0 & 54.55 {\footnotesize ($\pm$ 7.19)}\textsuperscript{\textdagger} \\ \hline
SQuAD & 0.2 & 62.91 {\footnotesize ($\pm$ 6.77)} \\ \hline
QQP & 0.5 & \textbf{76.36} {\footnotesize ($\pm$ 2.16)}* \\ \hline
\end{tabular}
\end{table}

\section{Related Work}

The importance of CR has been noted in prior ABSA work. \citet{ding-liu-2010-resolving} use aspect sentiments for performing CR, demonstrating a correlation between CR and sentiment classification. \citet{de-clercq-hoste-2020-absolutely} use CR to detect aspects from related reviews, for the reviews lacking explicit aspects. Instead, we consider an LLM's intra-sentence CR ability, considering only reviews with explicit aspects as having an aspect is critical to ALSC. \citet{mai-zhang-2020-aspect} use CR in aspect extraction, but only for identifying duplicate references among proposed aspects.\citet{varghese2013aspect} use CR to solve their dependency parser component's inability to correctly associate opinion words with pronouns. In our work, we consider the CR problem in end-to-end state-of-the-art ALSC LLM models. \citet{chen-etal-2020-knowledge} improve BERT LLM's CR ability for opinion-mining, using a method relying on external knowledge bases.

\section{Conclusion}

Since real world reviews vary widely, we need ALSC models which can handle various kinds of reviews, including those requiring CR. Although LLMs generally perform well on ALSC, our experiments provide evidence that LLMs can have poor performance on ALSC reviews requiring CR ability. We show that this problem can be alleviated by fine-tuning with certain auxiliary tasks before fine-tuning on the target tasks. Our framework for evaluating and improving an LLM's performance on CR cases can be applied for other tasks as well. Such a framework is critical for developing any model deployed in the real world. In the future, we will explore if auxiliary training can reduce the target task training that is needed for CR cases.

\section*{Limitations}
\begin{itemize}
    \item Even though we have successfully demonstrated a framework to handle CR-containing reviews by using auxiliary fine-tuning, we have not found which auxiliary tasks to definitively use for target tasks other than ALSC. The auxiliary task must be found using the framework proposed in our work.
    \item Our test set is composed of \textasciitilde 350 manually identified examples are guaranteed to require CR ability. However, it is common for ALSC datasets to be small. The bench-marking datasets Twitter, Lap14, Rest16 and Rest15 all have \textasciitilde 500-600 aspects for analysis \cite{zhang-etal-2019-aspect} which is close to our dataset. To reduce the variability due to a relatively small test set, we use multiple random seeds for robustness \cite{clark2020electra}. 
    
    Due to the specific problem we are targeting, it is difficult to create more examples than this using existing sources. During qualitative analysis, we had considered many ALSC datasets (SemEval datasets, Twitter, MAMS) but found that the CR problem was most pronounced in the restaurant domain (Rest16, MAMS). Example: laptop reviews rarely use explicit aspects \cite{pontiki-etal-2014-semeval}, leading to few CR cases in Lap14 dataset.
    \item Ours is the first work to demonstrate this CR problem in language models, thus there are few benchmarks against which we can compare our solution.
    \item We use the T5-large LLM for our experiments which requires a significant amount of computational resources for training. This leads to a high cost both financially and environmentally \cite{strubell-etal-2019-energy}.
\end{itemize}

\bibliography{anthology,custom}
\bibliographystyle{acl_natbib}

\newpage

\appendix


\section{Hyperparameters}

Learning rates for both auxiliary fine tuning and ALSC training steps are picked from $\{5e-4, 1e-4, 5e-5\}$ and $\{1e-3, 5e-4, 1e-4\}$ respectively, after running for three random seeds and selecting the rates giving max F1 score for their respective validation dataset. For auxiliary fine-tuning, the learning rates for all auxiliary tasks were found to be $1e-4$, except for SQuAD with Aux Fraction as 1.0 for which we found learning rate as $5e-5$. For ALSC target task training, the learning rate was found to be $5e-4$ in all cases except when using Commongen task for fine tuning with Aux Fraction as 0.1 for which we found learning rate as $1e-4$.

Batch size for training is taken as 16 to maximise GPU utilisation. We train for 30 epochs to allow for convergence, while using an early stopping mechanism.

\section{Model Details}

For our LLM, we use the T5-large implementation on Huggingface.\footnote{ \url{https://huggingface.co/t5-large}}

\section{Dataset Cleanup} \label{dataset_cleanup}

Following existing work \cite{tang-etal-2016-aspect, tian-etal-2021-aspect} we disregard reviews with no aspects, and also the aspects labeled as having "conflict" sentiment polarity to prevent a class imbalance problem due to low count of "conflict" class.

\section{Dataset Details} \label{app:dataset_details}

Here we present some more details of the ALSC-CR dataset. The aspect polarity distribution is presented in Table \ref{table:absa_cr_labels}. Note that it is possible to have multiple pronouns in each of the CR cases. 

The sentiment distribution of ALSC-CR test set is shown in Table \ref{table:absa_cr_test_pronouns}.

For constructing ALSC-CR, we use standard ALSC datasets (MAMS and Rest16). MAMS's original train set along with data from Rest16 train set is used for training. For validation, we use the original validation sets from MAMS and Rest16, in addition to Pronoun cases from MAMS test and Rest16. The composition of the validation dataset is such that we use minimal Pronoun cases for validation while having sufficient CR cases for testing. Details of the composition of ALSC-CR are shown in Table \ref{table:absa_cr_data_detailed}.


\begin{table}[!h]
\centering
\small
\caption{Sentiment polarity distribution in ALSC-CR dataset}
\label{table:absa_cr_labels}
\begin{tabular}{|l|c|c|c|}
\hline
\textbf{Partition} & \multicolumn{3}{c|}{\textbf{Polarity}} \\
& \textbf{Positive} & \textbf{Negative} & \textbf{Neutral} \\ \hline
Train & 4,279 & 3,065 & 5,090 \\ \hline
Validation & 337 & 222 & 330 \\ \hline
Test & 178 & 122 & 46 \\ \hline

\end{tabular}
\end{table}

\begin{table}[!h]
\centering
\small
\caption{Pronoun distribution in ALSC-CR test set, which has only CR cases}
\label{table:absa_cr_test_pronouns}
\begin{tabular}{|l|c|}
\hline
\textbf{Pronoun} & \textbf{Count} \\ \hline
it & 132 \\ \hline
which & 59 \\ \hline
they & 54 \\ \hline
he & 24 \\ \hline
who & 19 \\\hline
she & 17 \\ \hline
their & 14 \\ \hline
them & 12 \\ \hline
its & 10 \\ \hline
his & 10 \\ \hline
there & 10 \\ \hline
him & 5 \\ \hline
her & 5 \\ \hline
hers & 0 \\ \hline

\end{tabular}
\end{table}

\section{Error Analysis by Pronoun} \label{app:absa_cr_pronoun_ea}

We analyse the errors and improvements seen for individual pronouns (in reviews) when ALSC-CR is evaluated with different ALSC models. Since a few pronouns have very low counts as per Table \ref{table:absa_cr_test_pronouns}, we only analyse the ones which have count greater than 15. 

For all pronouns analysed, we find improvements in prediction accuracy for the models fine-tuned with auxiliary tasks, compared to the baseline model which has not auxiliary fine-tuning. Results are shown in Table \ref{table:absa_cr_pronoun_ea}.

\begin{table}[!h]
\centering
\small
\caption{Error Analysis of ALSC models by pronoun distribution. Model Accuracy\% presented by Pronoun. Highest scores bolded. $2^{nd}$ highest underlined. Pronouns with count less than 15 (as per Table \ref{table:absa_cr_test_pronouns}) are not analysed.}
\label{table:absa_cr_pronoun_ea}
\begin{tabular}{|l|c|c|c|}
\hline
\textbf{Pronoun} & \textbf{Baseline} & \textbf{Commongen 0.1} & \textbf{QQP 0.5} \\ \hline
it & 65.91 & \underline{68.18} & \textbf{71.21}\\ \hline
which & 74.58 & \textbf{83.05} & \underline{77.97}\\ \hline
they & 72.22 & \textbf{79.63} & \underline{77.78}\\ \hline
he & 70.83 & \textbf{75.0} & \underline{70.83}\\ \hline
who & 84.21 & \textbf{94.74} & \textbf{94.74}\\\hline
she & \underline{88.24} & \textbf{94.12} & \underline{88.24}\\ \hline
their & 64.29 & 78.57 & 78.57\\ \hline
them & 75.0 & 75.0 & 75.0\\ \hline
its & 80.0 & 70.0 & 90.0\\ \hline
his & 100.0 & 100.0 & 100.0\\ \hline
there & 60.0 & 70.0 & 60.0\\ \hline
him & 60.0 & 60.0 & 60.0\\ \hline
her & 100.0 & 100.0 & 80.0\\ \hline
hers & N/A & N/A & N/A\\ \hline

\end{tabular}

\end{table}

\section{Training Prompts} \label{app:training_prompts}

We present the training prompts used in Table \ref{table:training_prompts}.

\section{Visualising Auxiliary Training Results} \label{app:visual_absa_cr}

\begin{figure}[!]
\caption{Performance of ALSC models with aux training on ALSC-CR dataset.}
\label{fig:absa_cr_finetune}
\begin{tikzpicture}
\footnotesize
\begin{axis}[
    title={},
    xlabel={Auxiliary Dataset Fraction},
    ylabel={Mean F1},
    xmin=0, xmax=1.0,
    ymin=65, ymax=77.5,
    xtick={0,0.1,0.2,0.5,1.0},
    ytick={65,67.5,70,72.5,75,77.5},
    legend pos=south east,
    legend style={font=\tiny},
    ymajorgrids=true,
    grid style=dashed,
    every axis plot/.append style={thick},
]

\addplot+[
    color=blue, mark=*, dashed
    ]
    coordinates {
        (0.1,75.72) +- (0.5,1.14)
        (0.2,72.46) +- (0.5,2.21)
        (0.5,71.04) +- (0.5,3.50)
        (1.0, 71.45) +- (0.5,1.91)
    };

\addplot+[
    color=red, mark=*, 
    ]
    coordinates {
        (0.1,72.49) +- (0.5,2.79)
        (0.2,71.85) +- (0.5,2.98)
        (0.5,76.10) +- (0.5,1.26)
        (1.0,71.30) +- (0.5,2.19)
    };
    
\addplot+[
    color=olive, mark=*, dash dot
    ]
    coordinates {
        (0.1,71.79) +- (0.5,1.55)
        (0.2,71.45) +- (0.5,3.02)
        (0.5,72.60) +- (0.5,1.85)
        (1.0,73.12) +- (0.5,2.15)
    };
    
\addplot+[
    color=orange, mark=*, loosely dashed
    ]
    coordinates {
        (0.1,72.02) +- (0.5,1.88)
        (0.2,72.60) +- (0.5,2.07)
        (0.5,71.47) +- (0.5,3.24)
        (1.0,72.08) +- (0.5,2.25)
    };

\addplot[mark=dotted, black] {71.07};

\legend{Commongen, QQP, CosmosQA, SQuAD, Baseline (No Aux)}

\end{axis}
\end{tikzpicture}

\end{figure}
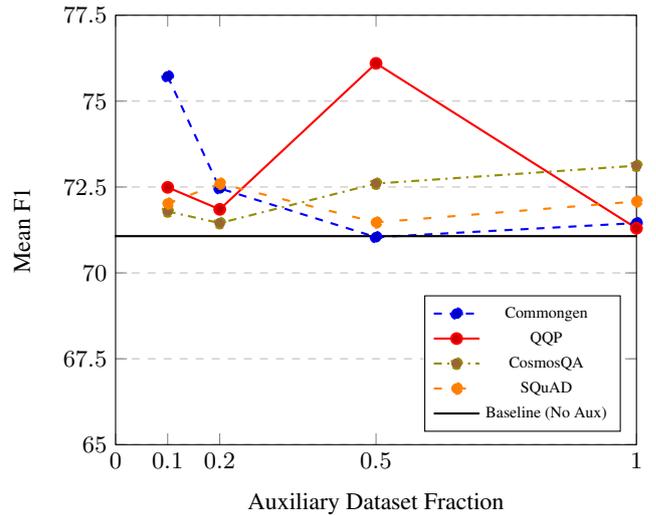

In Figure \ref{fig:absa_cr_finetune}, we visually show the performance of auxiliary trained models on ALSC-CR (same results as Table \ref{table:absa_cr_finetune}). We can see that there is little correlation between the auxiliary dataset fraction and the mean F1 performance, making it necessary to explore various fraction settings.

\section{Training Details}

For fine tuning the T5-large model, we use 1 NVIDIA V100 GPU, 6 CPU cores with 4 GB memory per core. We run training jobs with a 71 hour time limit.

\begin{table*}[!h]
\centering
\centering
\caption{Detailed ALSC-CR dataset composition.}
\label{table:absa_cr_data_detailed}
\small
\begin{tabular}{|l|l|l|}
\hline
\textbf{Partition} & \textbf{Size} & \textbf{Composition} \\ \hline
Train & 12,434 & \parbox{12cm}{\strut MAMS Train (\#count = 11,186) + Rest16 Train (Non Pronoun) (\#count = 1,248) \strut}\\ \hline
Val & 889 &  \parbox{12cm}{\strut 15\% of (MAMS Test (Pronoun) + Rest16 Train/Val/Test (Pronoun)) + 50\% of (MAMS Val + Rest Val (Non Pronoun)) [Here, MAMS \#count = 746, Rest16 \#count = 143]\strut} \\ \hline
Test & 346 & \parbox{12cm}{\strut MAMS Test (CR) (\#count = 124) + Rest16 Train/Val/Test (CR cases) (\#count = 222)}\\ \hline
\end{tabular}
\end{table*}

\begin{table*}[!h]
\caption{Details of T5 training prompts used for auxiliary and target tasks.}
\label{table:training_prompts}
\small
\begin{tabular}{|l|l|}
\hline
\textbf{Task} & \textbf{Training Prompt} \\ \hline
ALSC-CR & get sentiment: [sentence, aspect] \\ \hline
ALSC-Regular & get sentiment: [sentence, aspect] \\ \hline
DPR & Get antecedent: [sentence] \\ \hline
Commongen & generate a sentence with: [concepts] \\ \hline
CosmosQA & question: [question] answer\_0: [ans\_0] answer\_1: [ans\_1] answer\_2: [ans\_2] answer\_3: [ans\_3] context: [context] \\ \hline
SQuAD & question: [question] context: [context]\\ \hline
QQP & qqp question1: [question\_1] question2: [question\_2] \\ \hline
\end{tabular}
\end{table*}

\end{document}